\documentclass[letterpaper]{article}
\usepackage{main}
\usepackage{times}
\usepackage{helvet}
\usepackage{courier}
\usepackage[hyphens]{url}
\usepackage{graphicx}
\usepackage{natbib}
\usepackage{caption}
\usepackage{algorithm}
\usepackage{algorithmic}
\usepackage{adjustbox}
\usepackage{amsfonts}
\usepackage{amsmath}
\usepackage{tcolorbox}
\usepackage{booktabs}
\usepackage{multirow}
\usepackage{hyperref}
\usepackage{marvosym}

\nocopyright

\usepackage{newfloat}
\usepackage{listings}
\DeclareCaptionStyle{ruled}{labelfont=normalfont,labelsep=colon,strut=off}
\floatstyle{ruled}
\newfloat{listing}{tb}{lst}{}
\floatname{listing}{Listing}

\def\InfiR{{InfiAlign}}

\title{InfiAlign: A Scalable and Sample-Efficient Framework for Aligning LLMs to Enhance Reasoning Capabilities}
\author{
    Shuo Cai\textsuperscript{\rm 1},
    Su Lu\textsuperscript{\rm 2},
    Qi Zhou\textsuperscript{\rm 2},
    Kejing Yang\textsuperscript{\rm 1},
    Zhijie Sang\textsuperscript{\rm 1},
    Congkai Xie\textsuperscript{\rm 1},
    Hongxia Yang\thanks{Corresponding author.}\textsuperscript{\rm 1}\textsuperscript{\rm 2}
}
\affiliations{
    \textsuperscript{\rm 1}InfiX.ai
    \textsuperscript{\rm 2}The Hong Kong Polytechnic University\\
    \href{hongxia.yang@polyu.edu.hk}{hongxia.yang@polyu.edu.hk}
}

\usepackage{bibentry}

\begin{document}

\maketitle

\begin{abstract}

Large language models (LLMs) have exhibited impressive reasoning abilities on a wide range of complex tasks. However, enhancing these capabilities through post-training remains resource intensive, particularly in terms of data and computational cost. Although recent efforts have sought to improve sample efficiency through selective data curation, existing methods often rely on heuristic or task-specific strategies that hinder scalability. In this work, we introduce \textbf{InfiAlign}, a scalable and sample-efficient post-training framework that integrates supervised fine-tuning (SFT) with Direct Preference Optimization (DPO) to align LLMs for enhanced reasoning. At the core of InfiAlign is a robust data selection pipeline that automatically curates high-quality alignment data from open-source reasoning datasets using multidimensional quality metrics. This pipeline enables significant performance gains while drastically reducing data requirements and remains extensible to new data sources. When applied to the Qwen2.5-Math-7B-Base model, our SFT model achieves performance on par with DeepSeek-R1-Distill-Qwen-7B, while using only approximately 12\% of the training data, and demonstrates strong generalization across diverse reasoning tasks. Additional improvements are obtained through the application of DPO, with particularly notable gains in mathematical reasoning tasks. The model achieves an average improvement of 3.89\% on AIME 24/25 benchmarks. Our results highlight the effectiveness of combining principled data selection with full-stage post-training, offering a practical solution for aligning large reasoning models in a scalable and data-efficient manner. The model checkpoints are available at \href{https://huggingface.co/InfiX-ai/InfiAlign-Qwen-7B-SFT}{https://huggingface.co/InfiX-ai/InfiAlign-Qwen-7B-SFT}.

\end{abstract}

\section{Introduction}
\begin{figure*}[t]
\centering 
\includegraphics[width=1\textwidth]{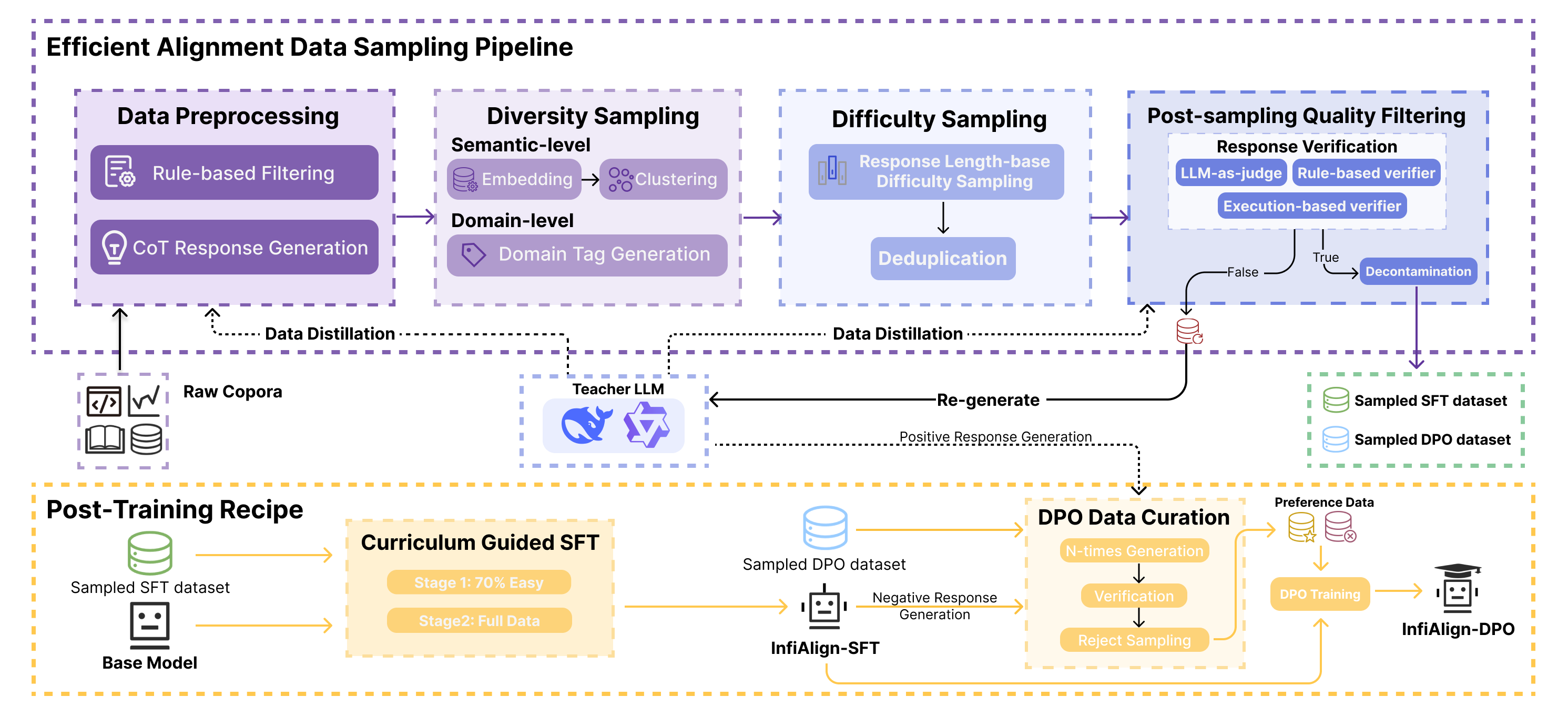}

\caption{\textbf{Overview of the InfiAlign Framework}  
The InfiAlign framework combines an efficient data sampling pipeline with a modular post-training strategy. The pipeline includes rule-based filtering, CoT distillation, diversity-aware sampling, and difficulty control via response length. The post-sampling quality filtering module applies both rule-based and LLM-based scoring. Post-training consists of a curriculum-guided SFT phase followed by a preference-based DPO stage. This framework enables scalable and automated generation of high-quality, domain-diverse alignment data.
}
\label{main-pipeline}
\end{figure*}

\noindent Large language models (LLMs) have demonstrated strong performance across a wide range of reasoning tasks, including mathematics, science, and programming. Post-training methods such as supervised fine-tuning (SFT) and reinforcement learning (RL)—often referred to as the alignment stage in LLM development—can further enhance reasoning capabilities, but they remain computationally expensive and data-intensive. These challenges are especially pronounced in chain-of-thought (CoT) reasoning \cite{wei2022chain}, which requires high-quality, domain-specific instruction data that are costly to curate and difficult to scale.

To address this, recent research has explored improving sample efficiency through selective data curation. Approaches such as model-based scoring \cite{chen2024alpagasustrainingbetteralpaca, ge2024clusteringrankingdiversitypreservedinstruction}, gradient-driven clustering \cite{zhang2024tagcostaskagnosticgradientclustered,xia2024lessselectinginfluentialdata,pan2024gdiggradientbaseddiversehighquality}, and embedding-based filtering \cite{bukharin2024datadiversitymattersrobust, wu2023selfevolveddiversedatasampling} have shown promising results. For domain-specific reasoning, multi-criteria selection methods like LIMO \cite{ye2025limoreasoning} and s1 \cite{muennighoff2025s1simpletesttimescaling} demonstrate that carefully curated, small-scale datasets—guided by factors such as difficulty, diversity, and generality—can yield substantial performance gains. Additionally, entropy-based compression techniques \cite{yin2024entropylawstorydata} aim to retain data diversity while reducing redundancy. However, many existing pipelines still suffer from critical limitations: they often rely on handcrafted heuristics (e.g., keyword filters or fixed scoring rules) or rigid teacher-student distillation schemes that lack generalization across tasks and domains \cite{li2024rule}. Moreover, these frameworks frequently require extensive manual effort or are tailored to specific domains, making them difficult to scale or adapt to new data sources. Such issues hinder the development of unified, automated, and broadly applicable alignment strategies for reasoning tasks.

In this work, we introduce \textbf{InfiAlign}, a unified and scalable post-training framework for aligning LLMs on reasoning tasks with high sample efficiency. InfiAlign integrates SFT and Direct Preference Optimization (DPO) \cite{rafailov2023direct}, built upon a robust data selection pipeline that automatically identifies high-quality alignment data from large open-source corpora using multi-dimensional metrics—capturing diversity, difficulty, and quality. Applied to the Qwen2.5-Math-7B-Base model, InfiAlign matches the performance of DeepSeek-R1-Distill-Qwen-7B while using only 20\% of the training data. Additional improvements are obtained through the application of DPO, with particularly notable gains in mathematical reasoning tasks. The model achieves an average improvement of 3.89\% on the AIME 2024 and AIME 2025 benchmarks. These results underscore the effectiveness of principled data selection and multi-stage alignment in enhancing LLM reasoning capabilities efficiently.

Our main contributions are as follows:

\begin{itemize}
  \item \textbf{Data-Efficient Alignment via Multi-Dimensional Filtering.}  
  We design an automated pipeline that selects high-quality instruction data from open-source corpora using diversity, difficulty, and quality metrics, achieving strong performance with only $\sim$20\% of the data used by distilled baselines.

  \item \textbf{Modular and Scalable Framework.}  
  InfiAlign enables seamless integration of new data sources and tasks via its modular design, allowing flexible and low-overhead adaptation across domains.

  \item \textbf{Enhanced Reasoning through Multi-Stage Training.}  
  We adopt a multi-stage training regimen that balances data mixing, curriculum-guided SFT, and DPO to boost reasoning across various benchmarks.
\end{itemize}

\section{Related Work}  

Recent advances in post-training have largely relied on data-intensive strategies such as SFT and RL to align LLMs for complex reasoning tasks. Many efforts construct reasoning datasets via distillation from stronger teacher models (e.g., QwQ \cite{qwq32b}, DeepSeek-R1 \cite{deepseekr1}), yielding models like DeepSeek-R1-Distill-Qwen and Light-R1 \cite{lightr1} that demonstrate strong downstream performance. However, these approaches often depend on heuristic or task-specific data collection pipelines, limiting their scalability and general applicability.

Several works (e.g., LIMO, s1) emphasize quality-over-quantity curation, showing that small yet carefully selected examples can be effective for reasoning supervision. Nonetheless, such efforts are either domain-specific or manually intensive, and do not scale well to broader alignment settings or new data sources. Beyond SFT, recent applications of DPO and other RL-based methods (e.g., AceReason \cite{chen2025acereason}, Skywork-OR1 \cite{he2025skywork}) further refine alignment, but do not prioritize generalizable data pipelines.

In contrast, our work proposes \textbf{InfiAlign}, a scalable and data-efficient post-training framework that integrates SFT and DPO under a unified and extensible data selection pipeline. By leveraging multi-dimensional quality metrics, our method enables high-quality alignment with minimal data, achieving competitive performance using substantially less training data compared to strong baselines. This framework provides a practical and generalizable foundation for future work on reasoning alignment, with potential to benefit researchers through more efficient model development at scale.

\section{InfiAlign: Scalable and Efficient Post-training for Reasoning}

We propose \textbf{InfiAlign}, a novel post-training framework that enhances the reasoning capabilities of large language models using minimal data. It integrates three core components: a \textbf{scalable data sampling pipeline} that efficiently selects a small yet high-quality subset of data by jointly considering diversity and difficulty, a \textbf{balanced SFT strategy} based on cross-domain data mixing for robust generalization, and a \textbf{data-efficient DPO recipe} that further strengthens reasoning capability. Together, these components enable strong performance with substantially reduced data and computational resources.

\subsection{Efficient Alignment Data Sampling Pipeline}

\noindent We introduce a scalable data pipeline for constructing high-quality QA pairs with controlled diversity and difficulty (see Figure~\ref{main-pipeline}). It consists of four components: (1) \textbf{Data Collection and Preprocessing}, which standardizes and optionally augments QA pairs with CoT reasoning; (2) \textbf{Diversity Sampling}, leveraging topic annotation and semantic clustering to ensure broad coverage; (3) \textbf{Difficulty Sampling}, which selects complex examples based on response characteristics; and (4) \textbf{Post-sampling Quality Filtering}, applying rule-based checks, sandbox verification, and LLM scoring. The resulting dataset is well-suited for alignment and distillation, especially for enhancing the reasoning abilities of small and medium language models.

\subsubsection{Data Collection and Preprocessing}

\noindent Alignment data is primarily collected from large-scale open-source reasoning datasets, with the flexibility to incorporate domain-specific or proprietary sources as needed. All data are formatted into QA pairs to support instruction alignment.
For queries lacking CoT reasoning traces, we generate responses using advanced models such as DeepSeek-Distill and Qwen3. Prior work demonstrates that such distillation effectively transfers reasoning abilities from larger models to smaller ones, enhancing alignment performance \cite{shridhar2022distilling, xu2024survey}.

We begin with rule-based filtering to remove non-English or incomplete QA pairs that may introduce noise. The filtered data are then processed by sampling modules to ensure broad coverage across query types and difficulty levels.

\subsubsection{Diversity Sampling}

\noindent To construct a high-quality alignment corpus that supports robust generalization and compositional reasoning, we introduce a dual-granularity diversity sampling strategy. This approach integrates both domain-level and semantic-level signals to capture topical breadth and latent linguistic diversity across QA instances.
\begin{itemize}

\item \textbf{Domain-Level Sampling:} We begin by assigning domain-specific labels to each question using a prompting-based LLM classifier (see Appendix~A). For structured domains such as mathematics and programming, we further decompose the hierarchy into fine-grained subcategories (e.g., \textit{Algebra}, \textit{Geometry}; \textit{Greedy Search}, \textit{Dynamic Programming}). 

Sampling is conducted in a category-balanced manner to avoid skewed distributions and promote balanced domain coverage.

\item \textbf{Semantic-Level Sampling.}
To promote diversity in the latent semantic space, we encode all questions into dense embeddings using a pretrained sentence encoder  (e.g., \texttt{Alibaba-NLP/gte-base-en-v1.5} \cite{zhang2024mgte}). We apply unsupervised clustering (e.g., K-means \cite{ahmed2020k}) over the embedding space and sample uniformly across clusters. This latent-space sampling strategy captures the variation in underlying semantics beyond the surface form, complementing domain-level sampling.

\end{itemize}
To finalize the candidate pool, sampling is performed independently at both levels and the results are merged. 

Deduplication is then applied using n-gram overlap matching ($n=20$), ensuring that samples containing common instructional templates shared across datasets are preserved and not erroneously discarded. Together, these complementary views synergistically enhance diversity across both topical and semantic dimensions, enabling the construction of heterogeneous and representative alignment corpora.

\begin{figure}[htbp]
\includegraphics[width=0.45\textwidth]{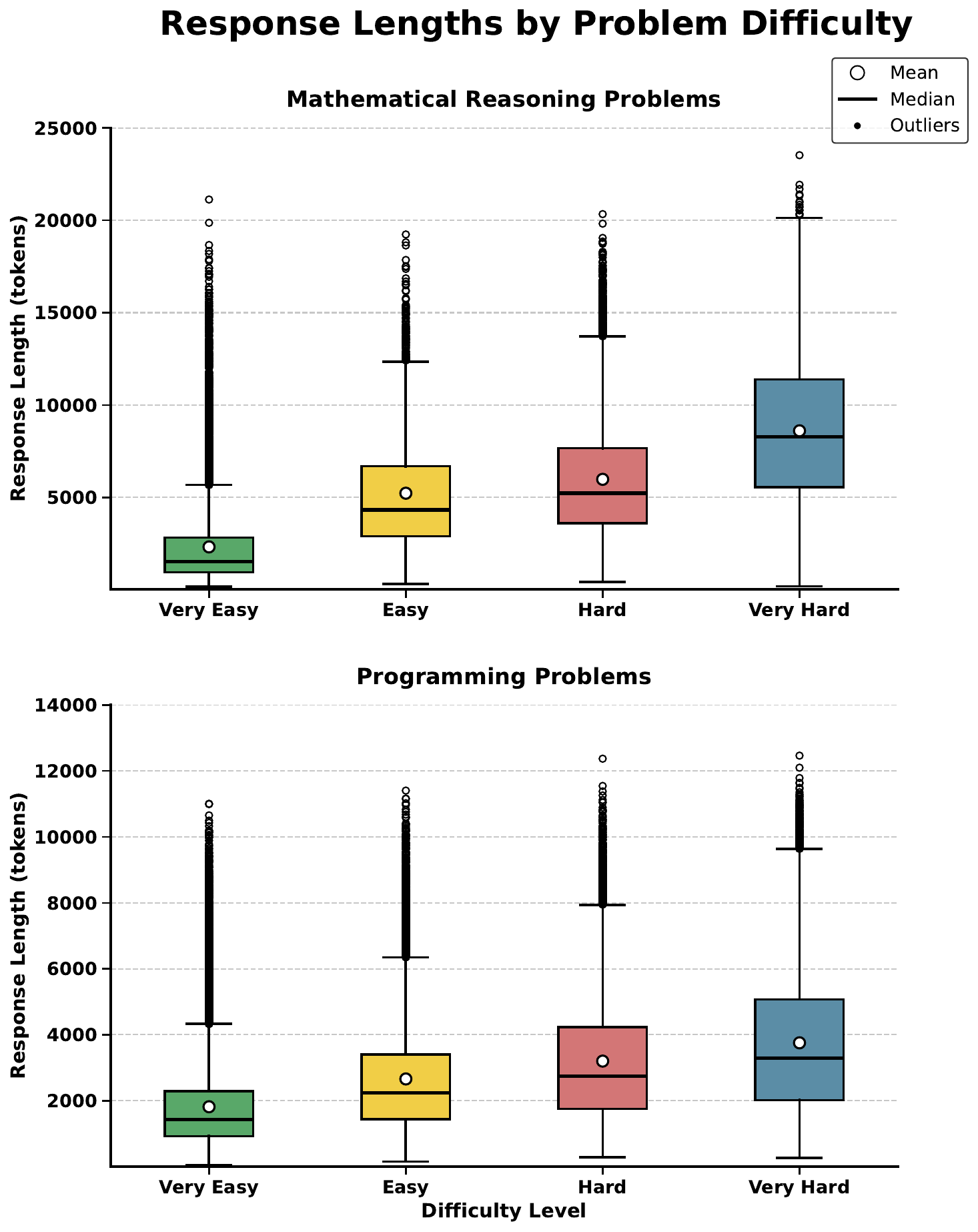}
\caption{Response lengths increase with problem difficulty across both mathematical and programming domains.
Box plots illustrate the distribution of response lengths (in tokens) across four difficulty levels for two problem categories. For both domains, higher difficulty is associated with longer responses, with mathematical problems exhibiting greater variance and heavier tails. This trend suggests that response length can serve as a coarse proxy for reasoning complexity in alignment data.} 
\label{fig:difficulty-corr}
\end{figure}

\subsubsection{Difficulty Sampling}

Figure~\ref{fig:difficulty-corr} presents a box plot of response lengths across four difficulty levels in math and programming tasks. Across both domains, we observe a clear positive correlation between task difficulty and model output length, consistent with the findings of OpenCodeReasoning \cite{ahmad2025opencodereasoning}. This empirical trend supports our use of response length as a scalable and domain-agnostic difficulty proxy. Unlike traditional pass@k-based difficulty estimation, which requires costly inference with oracle models, length-based sampling provides a practical alternative that generalizes across symbolic and semi-structured domains. In practice, we prioritize longer responses within each semantic or topical cluster, preserving both difficulty and diversity. This strategy improves reasoning power and generalization in downstream alignment, especially for complex tasks.

\subsubsection{Post-sampling Quality Filtering}

\noindent After sampling, we conduct a final quality control phase to ensure that only well-structured, accurate, and reliable QA pairs are retained for alignment training.

We begin with format-level validation to eliminate responses that are incomplete, excessively verbose, or missing critical components—such as final answers enclosed in \texttt{\textbackslash boxed{}} for mathematical problems. Domain-specific automated verifiers (e.g., \texttt{MathVerify}, \texttt{Sandbox}) are employed to assess response correctness in tasks with well-defined ground truth, such as mathematics and programming.
For responses that fail verification, we invoke an LLM to regenerate the answer using a structured correction template. This verification–regeneration process is iterated up to eight times or until all verification checks are passed. Responses that fail all attempts are discarded.

For open-ended or partially verifiable tasks, we employ LLM-based evaluation protocols to assess question clarity, answer redundancy, and overall informativeness. In cases where the response is ambiguous or the confidence is low, the sample is conservatively discarded to maintain the reliability of the dataset.

\subsubsection{Dataset Decontamination}

\noindent To avoid data leakage to evaluation benchmarks, we perform data decontamination. Specifically, we filter out QA pairs that exhibit substantial lexical or semantic overlap with publicly available benchmark datasets. This includes removing examples with high $n$-gram overlap ($n=15$) or elevated cosine similarity scores (greater than 0.9) based on sentence embeddings. This procedure helps prevent contamination of the test set and ensures that the evaluation metrics accurately reflect the generalization capabilities of the model.

These post-sampling quality control mechanisms, in conjunction with prior filtering and decontamination steps, ensure that the final alignment corpus is clean, diverse, and robust—suitable for high-quality instruction tuning.

\subsection{SFT Data Curation and Training Recipe}

\subsubsection{Data Sources and Composition}

\noindent To enable sample-efficient alignment via supervised fine-tuning, we curate \textbf{\InfiR-SFT-92K} and \textbf{\InfiR-SFT-165K}—two compact yet high-quality instruction corpus consisting of 95K or 165K reasoning-focused QA pairs. These datasets are constructed from over 10M raw alignment examples drawn from ten open-source corpora, including \texttt{OpenThoughts-114K}, \texttt{OpenThoughts3-1.2M} \cite{guha2025openthoughtsdatarecipesreasoning}, \texttt{AM-DeepSeek-R1-Distilled-1.4M} \cite{zhao202514millionopensourcedistilled}, \texttt{data-ablation-full59K} \cite{muennighoff2025s1simpletesttimescaling}, \texttt{NuminaMath-CoT} \cite{numina_math_datasets}, \texttt{OpenCodeReasoning}, \texttt{Llama-Nemotron-Post-Training-Dataset} \cite{bercovich2025llamanemotronefficientreasoningmodels}, \texttt{Mixture-of-Thoughts} \cite{mixtureofthoughts}, and \texttt{OpenScience} \cite{openscience}. Please refer to Appendix B for data composition and proportion.

To ensure that the resulting dataset is both informative and domain-balanced, we apply the proposed multi-dimensional data selection pipeline, which evaluates samples based on diversity, difficulty, and quality. Empirically, we observe that mathematical and coding tasks exhibit strong transferability and are more sensitive to data scaling, whereas general and domain-specific examples offer diminishing returns under increased volume. Based on these findings, we adopt a domain mixing ratio of \textbf{Math:Code:Science = 4:4:3}, prioritizing reasoning-rich tasks while maintaining a broad topical spread.

\subsubsection{Two-stage Curriculum Learning}

\noindent To further optimize learning dynamics and mitigate data inefficiency, we adopt a curriculum-inspired two-stage fine-tuning strategy that reflects the hierarchical complexity of reasoning tasks. In the first stage, we train the model on 70\% relatively simple data of the data (predominantly math and code instructions) which provide structured and relatively accessible reasoning patterns. This early phase allows the model to acquire foundational reasoning skills in a stable optimization regime.

In the second stage, we expand the training set to the full \InfiR-SFT-165K corpus by incorporating more diverse and domain-specific instructions, particularly from scientific and open-ended domains. Crucially, we retain first-stage samples in this phase to ensure distributional continuity and avoid catastrophic forgetting. This gradual curriculum enables the model to transition smoothly from well-structured to more open-ended reasoning tasks, leading to improved generalization across domains. Together, the domain-aware data composition and curriculum-based training schedule form a unified and principled strategy for effective reasoning alignment under limited data budgets.

\subsection{DPO Data Curation}

\noindent To further enhance the reasoning capability of our SFT model, we continue training it with DPO, one of the most popular preference optimization method. Given a prompt $x$ and a pair of responses $(y_w, y_l)$, where $y_w$ is the correct answer and $y_l$ is the SFT model's incorrect answer, DPO maximizes the log-likelihood gap between the correct answer and incorrect answer. The objective function of DPO is

\begin{equation}
\begin{aligned}
\mathcal{L}_{\mathrm{DPO}} = - \mathbb{E}_{(x, y_w, y_l) \sim \mathcal{D}} & \left[ \log \sigma \left( 
\beta \log \frac{\pi_\theta(y_w \mid x)}{\pi_{\mathrm{ref}}(y_w \mid x)} \right. \right. \\
&\quad \left. \left. - \beta \log \frac{\pi_\theta(y_l \mid x)}{\pi_{\mathrm{ref}}(y_l \mid x)}
\right) \right]
\end{aligned}
\end{equation}

\noindent where $\pi_\theta$ is the policy model, $\pi_{\mathrm{ref}}$ is the reference policy, typically the SFT model, $\sigma$ is the sigmoid function and $\beta$ controls the deviation from the base reference policy.

To build the DPO training dataset, we leverage \texttt{OpenMathReasoning} \cite{openmathreasoning}, \texttt{Mixture-of-Thoughts} and \texttt{OpenScience}, which provide QA pairs spanning the math, science and code domains. All samples in these datasets contain verified reasoning solutions generated by powerful reasoning models such as DeepSeek-R1 and QwQ-32B. The DPO data curation pipeline includes:

\begin{itemize}

\item \textbf{Data Decontamination and Deduplication:} We decontaminate data against evaluation benchmarks and deduplicate samples from the SFT training dataset.

\item \textbf{Data Selection:} We first utilize Qwen2.5-32B-Instruct model \cite{qwen2.5} to annotate each sample with domain-specific labels. For each category, we select the problems with the longest solution, representing the most challenging problems. Our SFT model then generates responses for these selected problems, which are used in the subsequent rejection sampling step.

\item \textbf{Reject Sampling:} We employ the Qwen2.5-32B-Instruct model to evaluate the SFT model's responses to math and science questions, and utilize an internal sandbox service to verify the correctness of code-related answers. For each domain, we select false samples with the longest solution lengths from each category, ensuring a balanced number of samples across categories. Previous work \cite{lightr1} has discovered that for challenging problems, using chosen responses from significantly stronger models yielded better results. Therefore, we directly use the solutions (generated by strong models such as DeepSeek-R1) as the positive samples, and pair them with the selected false samples to construct training pairs.

\end{itemize}

\begin{table*}[tbp]
\centering
\begin{adjustbox}{width=\linewidth}
\begin{tabular}{c|c|c|ccccccc}
\toprule
\multirow{2}{*}{\textbf{Model}} &\multirow{2}{*}{\textbf{Initial CKPT}} &\multirow{2}{*}{\textbf{Data Size}} & \textbf{AIME 2025} & \textbf{AIME 2024} & \textbf{MATH500} & \textbf{GPQA Diamond} & \textbf{MMLU-Pro} & \textbf{LCB-v5} & \multirow{2}{*}{\textbf{Avg.}} \\
 & & & (avg@64) & (avg@64) & (avg@4) & (avg@8) & (pass@1) & (avg@8) & \\

\midrule
Qwen2.5-7B-Instruct  & Qwen2.5-7B-Base & 1M & 8.80 & 11.93 & 76.15 & 38.70 & \textbf{57.49} & 15.77 & 34.80 \\
Qwen2.5-Math-7B-Instruct  & Qwen2.5-7B-Math-Base & 2.5M & 6.72 & 6.67 & 82.40 & 31.12 & 43.06 & 2.68 & 28.78 \\
DeepSeek-Distill-Qwen-7B & Qwen2.5-7B-Math-Base & 800K & 37.97 & 55.50* & 92.80* & 49.10* & 54.16 & 37.60* & 54.43 \\
OpenThinker2-7B & Qwen2.5-7B-Instruct & 1M & 38.70* & 60.70* & 87.60* & 47.00* & 40.60* & 37.50 & 52.01 \\
Light-R1-7B-DS & DeepSeek-Distill-Qwen-7B & 3K & 44.30* & 59.10* & 91.35 & 49.40* & 54.95 & \textbf{38.40} & 56.25 \\

\midrule
\InfiR-Qwen-7B-SFT-92K & Qwen2.5-7B-Math-Base & 92K & 43.39 & 56.46 & 92.35 & 48.48 & 53.51 & 34.05 & 54.70 \\
\InfiR-Qwen-7B-DPO-9K & \InfiR-Qwen-7B-SFT-92K & 9K & 44.06 & 61.04 & 91.95 & 48.17 & 49.90 & 34.54 & 54.94 \\

\midrule
\InfiR-Qwen-7B-SFT-165K & Qwen2.5-7B-Math-Base & 165K & 42.19 & \textbf{63.75} & 92.70 & \textbf{53.60} & 56.68 & 36.20 & \textbf{57.52} \\
\InfiR-Qwen-7B-DPO-10K & \InfiR-Qwen-7B-SFT-165K & 10K & \textbf{47.45} & 61.25 & \textbf{93.45} & 51.77 & 53.95 & 35.30 & 57.20 \\

\bottomrule
\end{tabular}
\end{adjustbox}
\caption{Main evaluation results of our \InfiR{} models on six representative reasoning benchmarks spanning mathematics, code, science, and general knowledge domains. All experiments are conducted under a unified evaluation setup (temperature=0.6, top\_p=0.95, max\_tokens=32{,}768). Results marked with * are self-reported by the model developers; the rest are reproduced using the same settings.}
\label{tab:main_eval_results}
\end{table*}

\section{Experiment}

\noindent We conduct comprehensive experiments to evaluate the effectiveness of our alignment data sampling pipeline in producing compact yet powerful instruction-tuned models. We first fine-tune a base model using supervised learning (\InfiR-SFT-7B), and further apply preference optimization (\InfiR-DPO-7B), both initialized from \texttt{Qwen2.5-Math-7B}.

Despite being trained on a relatively fewer alignment samples compared to other state-of-the-art models, both \InfiR-SFT-7B and \InfiR-DPO-7B demonstrate competitive or superior performance on general reasoning, math and code benchmarks.

\subsection{\InfiR-7B Training}

\noindent We use the datasets \InfiR-SFT-92k and \InfiR-SFT-165k to train \InfiR-Qwen-7B-SFT-92K and \InfiR-Qwen-7B-SFT-165K, respectively, which are constructed through our proposed aligned data sampling pipeline. We fine-tune \texttt{Qwen2.5-Math-7B} using a two-stage SFT schedule. The model is trained for 5 epochs using a batch size of 16 and a learning rate of 1e-5. All training is conducted on 8 NVIDIA H800 GPUs using mixed precision. The two-stage training first emphasizes simpler mathematical and code data before introducing more complex and general-domain examples, consistent with our curriculum-inspired strategy.

We conduct DPO training on both \InfiR-Qwen-7B-SFT-92K and \InfiR-Qwen-7B-SFT-165K models. To maintain the same data mixing strategy as used during SFT training, we construct two separate DPO training sets: \InfiR-DPO-9K (comprising 4k math, 3k code, and 2k science samples) for training \InfiR-Qwen-7B-SFT-92K model, and \InfiR-DPO-10K (comprising 3.5k math, 3.5k code, and 3k science samples) for training \InfiR-Qwen-7B-SFT-165K model.

We utilize 360-LLaMA-Factory framework \cite{360llamafactory} with sequence parallelism to train our DPO model on 16 NVIDIA H800 GPUs with the following settings: epoch as 3, batch size as 16, learning rate as 5e-7, cosine learning rate scheduler, warm-up ratio as 0.1, sequence parallelism as 4. Training minimizes the sigmoid preference loss with $\beta$ as 0.1.

\subsection{Evaluation}
\subsubsection{Benchmarks}
We evaluate our models on a diverse set of benchmarks covering four key domains: mathematical reasoning (AIME24/25~\citep{aime}, MATH500~\citep{math500}), code generation (LiveCodeBench~\citep{jain2024livecodebench}), general reasoning (MMLU-Pro~\citep{mmlupro}), and scientific QA (GPQA-Diamond~\citep{team2025supergpqa}). This suite provides a comprehensive evaluation of both domain-specific and general instruction-following capabilities.

\subsubsection{Baselines}
\noindent We compare our approach against multiple strong reasoning baselines, including DeepSeek-Distill-Qwen-7B, OpenThoughts2-7B \cite{guha2025openthoughts}, and Light-R1-7B-DS \cite{lightr1}. These models are either trained on substantially larger datasets or built upon more powerful base models.

During evaluation, we use a sampling temperature of 0.6 and top-p of 0.95 across all benchmarks. The maximum generation length is set to 32,768 tokens for all tasks. To address variability in reasoning outputs, we report pass@1 performance averaged over multiple runs (denoted as avg@n): \(n=64\) for AIME 24/25, \(n=4\) for MATH500, \(n=8\) for GPQA-Diamond, LiveCodeBench, and \(n=1\) for MMLU-Pro.

\subsection{Main Results}

Table~\ref{tab:main_eval_results} presents the performance of our models across six representative reasoning benchmarks. \InfiR-Qwen-7B-SFT-92K achieves an average accuracy of 54.70, matching or slightly exceeding \texttt{DeepSeek-Distill-Qwen-7B} (54.43) while using only \textbf{12\%} of the training data (92K vs.\ 800K). Notably, it generalizes well to both mathematical (AIME 2025: 43.39 vs.\ 38.70) and scientific domains (GPQA: 48.48 vs.\ 47.00), outperforming several baselines trained on substantially larger datasets or with stronger backbones (e.g., OpenThoughts2-7B). These results highlight the effectiveness of our sample-efficient alignment pipeline in achieving strong reasoning generalization under minimal supervision.

To evaluate scalability, we further apply the same sampling pipeline to scale up the training set to 165K QA pairs. The resulting model, \InfiR-Qwen-7B-SFT-165K, achieves a higher average accuracy of 57.52, with consistent improvements over the 92K variant across most benchmarks—including +7.29 on AIME 2024, +5.12 on GPQA, and +2.15 on LCB-v5. This upward trend underscores the robustness and scalability of our method, allowing practitioners to balance training cost and performance based on resource availability.

Finally, lightweight preference tuning via DPO could further boost math reasoning ability. On math domain benchmarks, compared to their respective SFT baselines, \InfiR-Qwen-7B-DPO-9K and \InfiR-Qwen-7B-DPO-10K achieve average improvements of 1.62\% and 1.18\%, respectively. Specifically, \InfiR-Qwen-7B-DPO-9K improves the AIME 2024 score with a +4.58 gain (61.04 vs. 56.46). While \InfiR-Qwen-7B-DPO-10K achieves 47.45 (+5.26) on AIME 2025 and 93.45 on MATH500, outperforming all baseline models. This highlights the complementary benefits of minimal yet targeted preference data in enhancing reasoning alignment.

\subsection{Ablation Studies and Analysis}

\subsubsection{Ablation Studies on Data Sampling Strategy}

\noindent  In this section, we conduct ablation studies to evaluate the impact of different data sampling strategies on the alignment performance of our model. To facilitate reproducibility, we set the random seed to 42 for all experiments.

\textbf{Effectiveness on General Reasoning.}
To evaluate the impact of general domain sampling strategies on alignment performance, we performed ablation experiments using fixed subsets of 17.1K QA pairs sampled from the AM-1.4M dataset~\cite{zhao202514millionopensourcedistilled}. Models trained on these subsets are evaluated on four representative benchmarks: MATH500, GPQA-Diamond, MMLU-Pro, and the more comprehensive SuperGPQA~\cite{team2025supergpqa}.

Table~\ref{tab:general_sampling_ablation} presents a comparison of eight sampling strategies, including random selection, length- and complexity-based filtering, and combinations with diversity mechanisms such as semantic-level embeddings, domain-level categorization, and our proposed dual-granularity approach.

\begin{table}[ht]
\centering
\begin{adjustbox}{width=\columnwidth}
\begin{tabular}{lcccc}
\toprule
\textbf{Strategy} & \textbf{MATH500} & \textbf{GPQA-Diamond} & \textbf{SuperGPQA} & \textbf{MMLU-Pro} \\
\midrule
Random & 75.60 & 33.21 & 22.96 & 50.31 \\
Dual diverse only & 76.25 & 32.13 & 23.43 & 48.49 \\
Length only & 83.30 & 35.81 & 30.02 & \textbf{56.88} \\
Complexity \& Dual diverse & 73.55 & \textbf{42.17} & 24.54 & 50.15 \\
IFD \& Dual diverse & 78.85 & 33.65 & 26.07 & 52.55 \\
Length \& Embedding diverse & \textbf{84.55} & 40.91 & 29.84 & 55.07 \\
Length \& Category diverse & 81.25 & 31.12 & 28.74 & 53.29 \\
\textbf{Length \& Dual diverse (Ours)} & 82.21 & 37.82 & \textbf{30.20} & \textbf{55.13} \\
\bottomrule
\end{tabular}
\end{adjustbox}
\caption{Ablation study on general data sampling strategies. Each strategy samples 17.1K instances from AM-1.4M. SFT was performed on the Qwen2.5-7B-Base model.}
\label{tab:general_sampling_ablation}
\end{table}

Sampling based on \textbf{response length} exhibits a strong correlation with enhanced mathematical reasoning, yielding a +7.7 point improvement over random sampling on MATH500 and outperforming all other diversity-driven strategies. This confirms response length as a reliable and efficient proxy for reasoning complexity in symbolic domains. In contrast, \textbf{complexity-aware sampling}—guided by model-estimated prompt difficulty—achieves superior performance on scientific tasks such as GPQA-Diamond, effectively capturing nuanced, knowledge-intensive challenges that length alone fails to reflect.

Regarding \textbf{diversity}, our \textbf{Length \& Dual diverse} approach, which integrates response-length heuristics with both domain-level and semantic-level diversity, consistently delivers balanced gains across all benchmarks. It achieves top performance on SuperGPQA and remains competitive elsewhere, outperforming single-axis diversity strategies (Length \& Embedding or Length \& Category). This underscores the importance of hybrid multi-granularity diversity in covering the heterogeneity of real-world instruction distributions. 

Collectively, these findings validate the central hypothesis of our framework: that a simple yet principled combination of response length and principled diversity is sufficient to construct compact, high-quality reasoning datasets. In contrast to approaches that rely on expensive difficulty estimators or task-specific heuristics, our method is lightweight, domain-agnostic, and empirically robust across diverse reasoning benchmarks.

\textbf{Effectiveness on Science and Math Reasoning.}
To further evaluate the domain-specific utility of our sampling strategy, we conduct ablation studies in two reasoning-intensive domains: \textbf{science} and \textbf{math}. For science, we sample 10K instances from the \texttt{OpenScience} dataset; for math, we consider \texttt{NuminaMath-CoT}, \texttt{s1-59K}, and their mixture. All models are fine-tuned from the \texttt{Qwen2.5-7B-Base} checkpoint under consistent settings.

\begin{table}[ht]
\centering
\begin{adjustbox}{width=\columnwidth}
\begin{tabular}{lccc}
\toprule
\textbf{Strategy (Science)} & \textbf{MATH500} & \textbf{GPQA} & \textbf{MMLU-Pro} \\
\midrule
Random & \textbf{72.00} & 40.23 & 53.10 \\
Dual diverse only & \textbf{70.05} & \textbf{42.74} & 53.70 \\
Length only & 65.90 & 38.51 & \textbf{57.35} \\
\textbf{Length \& Dual diverse (Ours)} & 69.95 & \textbf{41.28} & \textbf{56.94} \\
\bottomrule
\end{tabular}
\end{adjustbox}
\caption{Ablation on science-domain sampling strategies. Each subset is drawn from \texttt{OpenScience}. 10k samples are used for each group.}
\label{tab:science_sampling_ablation}
\end{table}

\begin{table}[ht]
\centering
\begin{adjustbox}{width=\columnwidth}
\begin{tabular}{lccccc}
\toprule
\textbf{Strategy (Math)} & \textbf{AIME25} & \textbf{AIME24} & \textbf{MATH500} & \textbf{GPQA} \\
\midrule
NuminaMath-CoT Easy\&diverse & 14.27 & 12.97 & 73.25 & 27.97 \\
NuminaMath-CoT Hard\&diverse & 20.57 & 15.05 & 76.45 & 33.90 \\
s1-59K Hard\&diverse & \textbf{23.85} & 22.71 & \textbf{84.40} & 32.51 \\
Mix Hard\&diverse & 21.72 & \textbf{26.82} & 82.00 & \textbf{35.54} \\
\bottomrule
\end{tabular}
\end{adjustbox}
\caption{Ablation on math-domain sampling. “Easy”/“Hard” defined by response length. “Mix” combines \texttt{NuminaMath-CoT} and \texttt{s1-59K}. Each group has 10K samples.}
\label{tab:math_sampling_ablation}
\end{table}

In the \textit{science domain}, unlike general data, diversity is a more critical factor due to the unique characteristics of different scientific subfields. Although Dual Diverse achieves only a slightly higher score on GPQA than our method, our Length \& Dual Diverse approach consistently yields balanced performance across other benchmarks (Table~\ref{tab:science_sampling_ablation}).

As shown in Table~\ref{tab:math_sampling_ablation}, performance in the \textit{math domain} improves with both data quality and instance difficulty. Longer, more diverse samples from \texttt{NuminaMath-CoT} outperform shorter ones, with notable gains on AIME25 (+6.3\%) and GPQA (+5.9\%). Samples drawn from \texttt{s1-59K} further exceed those from \texttt{NuminaMath-CoT} alone, indicating higher source quality. Importantly, combining both sources using our dual-heuristic strategy achieves the best overall results, highlighting the approach’s robustness and scalability in multi-source alignment settings.

\subsubsection{Scaling InfiAlign to 32B: Robustness Across Model Sizes}

\noindent We evaluate the scalability of \textbf{InfiAlign} beyond 7B by fine-tuning \textbf{Qwen2.5-32B-Instruct} on 1K-sample subsets drawn from a shared 59K data pool, with strict de-duplication via 15-gram filtering and embedding similarity (\textgreater 0.9). All responses are generated by a high-capacity teacher model (QwQ-32B) and evaluated on four reasoning benchmarks.

\begin{table}[ht]
\centering

\begin{adjustbox}{width=\columnwidth}
\begin{tabular}{lccccc}
\toprule
\textbf{Data} & \textbf{AIME 2025} & \textbf{AIME 2024} & \textbf{MATH500} & \textbf{GPQA Diamond} & \textbf{Avg.} \\
\midrule
s1.1 & 56.70 & 60.00 & \textbf{95.40} & 63.60 & 68.93 \\
\midrule
s1K-QwQ & 59.38 & \textbf{67.29} & 94.35 & \textbf{67.31} & 72.08 \\
Random 1k & 55.78 & 63.44 & 93.75 & 64.52 & 69.37 \\
Random 1k & 56.15 & 64.53 & 94.00 & 66.60 & 70.32 \\
Random 1k & 57.55 & 65.05 & 94.30 & 64.65 & 70.39 \\
InfiAlign 1k & \textbf{63.70} & 66.61 & 94.75 & 64.07 & \textbf{72.28} \\
\bottomrule
\end{tabular}
\end{adjustbox}
\caption{Ablation study on Qwen2.5-32B-Instruct using different 1k-sample subsets from the same 59K data pool. The responses of all samples were generated using QwQ-32B. }
\label{tab:ablation32b}
\end{table}

\noindent Table~\ref{tab:ablation32b} reveals key insights:

\textbf{High-quality supervision is crucial.} Replacing DeepSeek-R1 with QwQ-32B supervision consistently improves s1K-QwQ over s1.1 across all benchmarks, notably +7.29 on AIME 2024. Linguistic analysis of reasoning-related discourse markers—such as deliberation cues (“let me think,” “hmm”), verification phrases (“let me double-check”), and supplemental expressions (“for example,” “on the other hand”)—shows that QwQ-32B responses are on average 20\% longer and contain 78\% more reasoning-indicative phrases (Table~\ref{tab:reasoning_features}). This suggests richer, more structured reasoning aligned with our hypothesis that longer responses encode stronger introspective signals, enhancing downstream distillation.

\begin{table}[ht]
\centering
\begin{adjustbox}{width=\columnwidth}
\begin{tabular}{lcc}
\toprule
\textbf{Teacher Model} & \textbf{Avg. Length (chars)} & \textbf{Total Feature Frequency} \\
\midrule
s1.1 (DeepSeek-R1) & 27,535.5 & 75.83 \\
s1K-QwQ (QwQ-32B) & \textbf{33,053.3(+20\%)} & \textbf{135.26(+78\%)} \\
\bottomrule
\end{tabular}
\end{adjustbox}
\caption{Linguistic characteristics of 1k responses generated by different teacher models.}
\label{tab:reasoning_features}
\end{table}

\textbf{InfiAlign demonstrates robustness and scalability.} It matches s1K-QwQ performance without task-specific heuristics and consistently outperforms random baselines. Compared to the manual, resource-intensive filtering in s1, our automated pipeline offers a scalable, generalizable solution across model sizes and domains. These results underscore the effectiveness of combining scalable quality assessment with principled data sampling to build high-performance alignment models.

\section{Conclusion and Limitation}
We propose \textbf{InfiAlign}, a scalable and data-efficient post-training framework that combines supervised fine-tuning and reinforcement learning to align large language models for complex reasoning tasks. Central to our approach is a robust data selection pipeline that leverages multi-dimensional quality metrics—diversity, difficulty, and alignment quality—to automatically curate high-value instruction data from open sources. Applied to Qwen2.5-Math-7B-Base, InfiAlign matches the performance of DeepSeek-R1-Distill-Qwen-7B while using only $\sim$12\% of the training data. Incorporating DPO further improves mathematical reasoning ability, with an 3.89\% average gain on AIME 24/25. The modularity of our pipeline allows for seamless integration of new tasks and data sources, supporting efficient scaling and continuous improvement.

\textbf{Limitations.} Although our selection framework is domain-agnostic, it relies on manually defined metrics that may require tuning for unseen domains. Furthermore, while response length and reasoning-indicative markers are positively correlated with model performance, we have not yet systematically investigated how these surface-level characteristics—particularly response diversity and linguistic markers—impact the effectiveness of student model distillation.

\bibliography{main}

\appendix
\section*{Appendix A: Domain Classification Prompts}
\label{sec:domain_classification_prompts}
We use prompting-based LLM classification to annotate each QA pair with fine-grained domain labels. Below, we present the exact prompts used for data annotation, including category descriptions for mathematics, code, science, and general instruction tasks.

\onecolumn
\noindent\textbf{Math Domain Classification Prompt}
\begin{tcolorbox}[title=Mathematics Domain Annotation Prompt, 
                  colback=gray!5!white, 
                  colframe=gray!75!black, 
                  width=\textwidth]

You are a mathematics expert tasked with classifying math problems into the correct categories:1. **Algebra**: Problems involving equations, expressions, polynomials, factorization, and number manipulation.
2. **Geometry and Topology**: Questions about shapes, angles, distances, volumes, or geometric relationships.
3. **Analysis**: Problems involving limits, calculus, differential equations, integrals, or series.
4. **Number Theory**: Focused on divisibility, primes, modular arithmetic, factorials, and integer properties.
5. **Probability and Statistics**: Questions involving likelihood, counting, distributions, expected values, or averages.
6. **Discrete Mathematics and Combinatorics**: Problems involving counting, arrangements, graphs, or logical structures.
7. **Logic**: Problems involving formal reasoning, truth tables, proof systems, predicates, and logical structure.

According to the {item['instruction']},
DO NOT PROVIDE ANY EXPLANATION.
JUST output ONLY THE MOST RELEVANT category name from the list above.
Put the most concise answer inside \texttt{\textbackslash boxed\{\}}.
\end{tcolorbox}

\noindent\textbf{Code Domain Classification Prompt}
\begin{tcolorbox}[title=Code Domain Annotation Prompt, colback=gray!5!white, colframe=gray!75!black, width=\linewidth]
You are a programming expert tasked with classifying coding problems into the appropriate category:1. **Array **: Problems focused on manipulating and traversing arrays, including operations like insertion, deletion, and searching within linear data structures.
2. ** String **: Challenges involving text processing, such as substring manipulation, pattern matching, and character encoding.
3. ** Hash Table **: Tasks leveraging key-value pair structures for efficient data lookup, insertion, or frequency counting.
4. ** Dynamic Programming **: Optimization problems requiring breaking down into overlapping subproblems and storing intermediate results.
5. ** Math **: Problems solvable using mathematical concepts, including arithmetic, algebra, or combinatorics.
6. ** Sorting **: Algorithms to reorder data based on specific criteria, often as a preprocessing step for other operations.
7. ** Greedy **: Problems where locally optimal choices at each step lead to a globally optimal solution.
8. ** Depth-First Search (DFS) **: Tree or graph traversal exploring as far as possible along each branch before backtracking.
9. ** Binary Search **: Efficient search algorithm for sorted datasets by repeatedly dividing the search interval in half.
10. ** Matrix **: Operations on 2D grids, such as traversal, rotation, or element-wise computations.
11. ** Bit Manipulation **: Problems requiring operations at the bit level, like masking or shifting.
12. ** Breadth-First Search (BFS) **: Level-order traversal for trees or graphs, often used to find shortest paths.
13. ** Two Pointers **: Technique using paired references to traverse data structures, often for pairwise comparisons or windowing.
14. ** Tree **: Hierarchical data structure problems involving nodes, paths, or subtree properties.
15. ** Prefix Sum: Preprocessing arrays to enable rapid range sum queries or cumulative calculations.
16. ** Heap (Priority Queue) **: Problems requiring efficient access to the highest/lowest priority element, often for scheduling.
17. ** Simulation **: Step-by-step emulation of real-world processes or system behaviors.
18. ** Stack **: Last-in-first-out (LIFO) structure problems, such as parsing or backtracking.
19. ** Counting **: Frequency analysis of elements, often combined with hash tables or arrays.
20. ** Graph **: Problems involving nodes and edges, such as pathfinding or cycle detection.
21. ** Sliding Window **: Technique to maintain a dynamic subset of data (e.g., contiguous subarrays) for efficiency.
22. ** Backtracking **: Exhaustive search by incrementally building candidates and abandoning invalid paths.
23. ** Enumeration **: Systematic listing of all possible solutions or configurations.
24. ** Union Find **: Disjoint-set operations for dynamic connectivity and component merging.
25. ** Monotonic Stack **: Stack variant maintaining elements in sorted order for next-greater/smaller problems.
26. ** Number Theory **: Mathematical problems focusing on integers, primes, divisibility, or modular arithmetic.
27. ** Linked List **: Linear data structure problems involving node manipulation and pointer traversal.
28. ** Bitmask **: Compact representation of sets or states using binary digits, enabling efficient bitwise operations.
29. ** Divide and Conquer **: Problems split into independent subproblems, solved recursively (e.g., merge sort).
30. ** Trie **: Tree-like structure for efficient string storage and retrieval (e.g., autocomplete).
31. ** Memoization **: Optimization technique caching intermediate results to avoid redundant computations.
32. ** Ordered Set **: Data structure maintaining sorted elements for rank/range queries.
33. ** Recursion **: Function calling itself to solve problems with repetitive substructures (e.g., Fibonacci).

According to the {item['instruction']},
DO NOT PROVIDE ANY EXPLANATION.
JUST output ONLY THE MOST RELEVANT category name from the list above.
Put the most concise answer inside \texttt{\textbackslash boxed\{\}}.
\end{tcolorbox}

\newpage
\noindent\textbf{Science Domain Classification Prompt}
\begin{tcolorbox}[title=Science Domain Annotation Prompt, colback=gray!5!white, colframe=gray!75!black, width=\linewidth]
You are a science expert tasked with classifying this question into a specific science discipline:1. **Molecular Biology**: Studies molecular processes in cells, such as DNA transcription, RNA translation, and protein synthesis.2. **Genetics**: Focuses on heredity and genetic variation, including gene inheritance, mutations, and trait transmission.3. **Other Biology**: Covers all other biological topics, such as cell biology, physiology, neuroscience, evolution, and ecology.4. **Quantum Mechanics**: Explores matter and energy behavior at atomic and subatomic scales, including superposition and entanglement.5. **High-Energy Particle Physics**: Investigates fundamental particles and forces using particle accelerators and the Standard Model.6. **Physics (general)**: Covers broad or foundational topics in classical or modern physics, not tied to a specific subfield.7. **Astrophysics**: Applies physical principles to celestial phenomena like stars, galaxies, and black holes.8. **Electromagnetism and Photonics**: Studies electric/magnetic fields, electromagnetic waves, light, and optics technologies like lasers.9. **Relativistic Mechanics**: Examines motion and gravity under Einstein's relativity, especially at near-light speeds.10. **Statistical Mechanics**: Uses probability to connect microscopic particle behavior to macroscopic physical laws like thermodynamics.11. **Condensed Matter Physics**: Studies the physical properties of solids and liquids, including semiconductors and superconductors.12. **Optics and Acoustics**: Examines the behavior of light and sound, including reflection, diffraction, and wave propagation.13. **Organic Chemistry**: Focuses on the structure and reactions of carbon-containing compounds, including biomolecules.14. **Chemistry (general)**: Covers fundamental chemical principles not limited to any specific subfield.15. **Inorganic Chemistry**: Studies compounds without carbon-hydrogen bonds, including metals and minerals.16. **Analytical Chemistry**: Involves methods to detect and quantify substances, such as spectroscopy and titration.17. **Physical Chemistry**: Combines physics and chemistry to study energy, thermodynamics, kinetics, and molecular behavior.18. **Others**: Use this only if the content clearly does not fit any of the categories above.

According to the {item['instruction']},
DO NOT PROVIDE ANY EXPLANATION.
JUST output ONLY THE MOST RELEVANT category name from the list above.
Put the most concise answer inside \texttt{\textbackslash boxed\{\}}.
\end{tcolorbox}

\newpage
\noindent\textbf{General Instruction Classification Prompt}
\begin{tcolorbox}[title=General Domain Annotation Prompt, colback=gray!5!white, colframe=gray!75!black, width=\linewidth]
You are a classification assistant tasked with identifying the general topic of a given instruction. Choose from the following categories:1. **Logic and Reasoning**: Problems involving deductive reasoning, logical puzzles, critical thinking, and formal systems.
2. **Mathematical Ability**: Questions requiring numerical computation, problem-solving, mathematical concepts, and quantitative analysis.
3. **Programming and Software Development**: Challenges related to coding, algorithms, data structures, debugging, and software engineering.
4. **Natural Language Processing and Understanding**: Tasks involving text analysis, language modeling, syntax, semantics, and machine comprehension.
5. **Information Processing and Integration**: Problems about data organization, knowledge synthesis, and combining multiple sources of information.
6. **Problem Solving and Support**: Questions that require troubleshooting, decision-making, and providing solutions to user queries.
7. **Data Science and Analytics**: Challenges involving statistical analysis, data visualization, predictive modeling, and machine learning.
8. **Creativity and Design**: Tasks related to ideation, artistic expression, UX/UI design, and innovative problem-solving.
9. **STEM Knowledge**: Interdisciplinary STEM applications (e.g., physics problem-solving, engineering principles)
10. **Humanities, History, Philosophy, and Sociology Knowledge**: Topics related to cultural studies, historical events, ethical theories, and social dynamics.
11. **Open Knowledge Q\&A**: General factual queries spanning a wide range of subjects without a specific domain focus.
12. **Life Knowledge and Skills**: Practical advice on everyday activities, DIY tasks, personal development, and lifestyle tips.
13. **Education and Consulting**: Questions about learning strategies, academic guidance, tutoring, and professional advice.
14. **Linguistic Knowledge, Multilingual and Multicultural Understanding**: Topics involving languages, translation, cultural nuances, and comparative linguistics.
15. **Medical, Pharmaceutical, and Health Knowledge**: Questions related to diseases, treatments, pharmacology, and wellness.
16. **Communication and Social Media**: Challenges involving interpersonal skills, digital marketing, content creation, and online engagement.
17. **Task Generation**: Requests for generating structured tasks, prompts, or problem sets for various applications.
18. **Literary Creation and Artistic Knowledge**: Topics related to writing, poetry, storytelling, visual arts, and creative expression.
19. **Analysis and Research**: Tasks requiring deep investigation, literature review, experimental design, and evidence synthesis.
20. ** Project and Task Management**: Questions about planning, organization, workflow optimization, and productivity strategies.
21. **Financial, Economic, and Business Knowledge**: Topics covering investments, market trends, accounting, and corporate strategies.
22. **Psychological Knowledge**: Questions related to cognitive processes, mental health, behavioral theories, and emotional intelligence.
23. **Open Task Completion**: Miscellaneous requests that don’t fit neatly into other categories but require structured execution.

According to the {item['instruction']},
DO NOT PROVIDE ANY EXPLANATION.
JUST output ONLY THE MOST RELEVANT category name from the list above.
Put the most concise answer inside \texttt{\textbackslash boxed\{\}}.
\end{tcolorbox}
\newpage
\section*{Appendix B: Data Composition and Proportion}
\label{sec:data_composition_and_proportion}
\begin{figure*}[h]
\centering 
\includegraphics[width=1\textwidth]{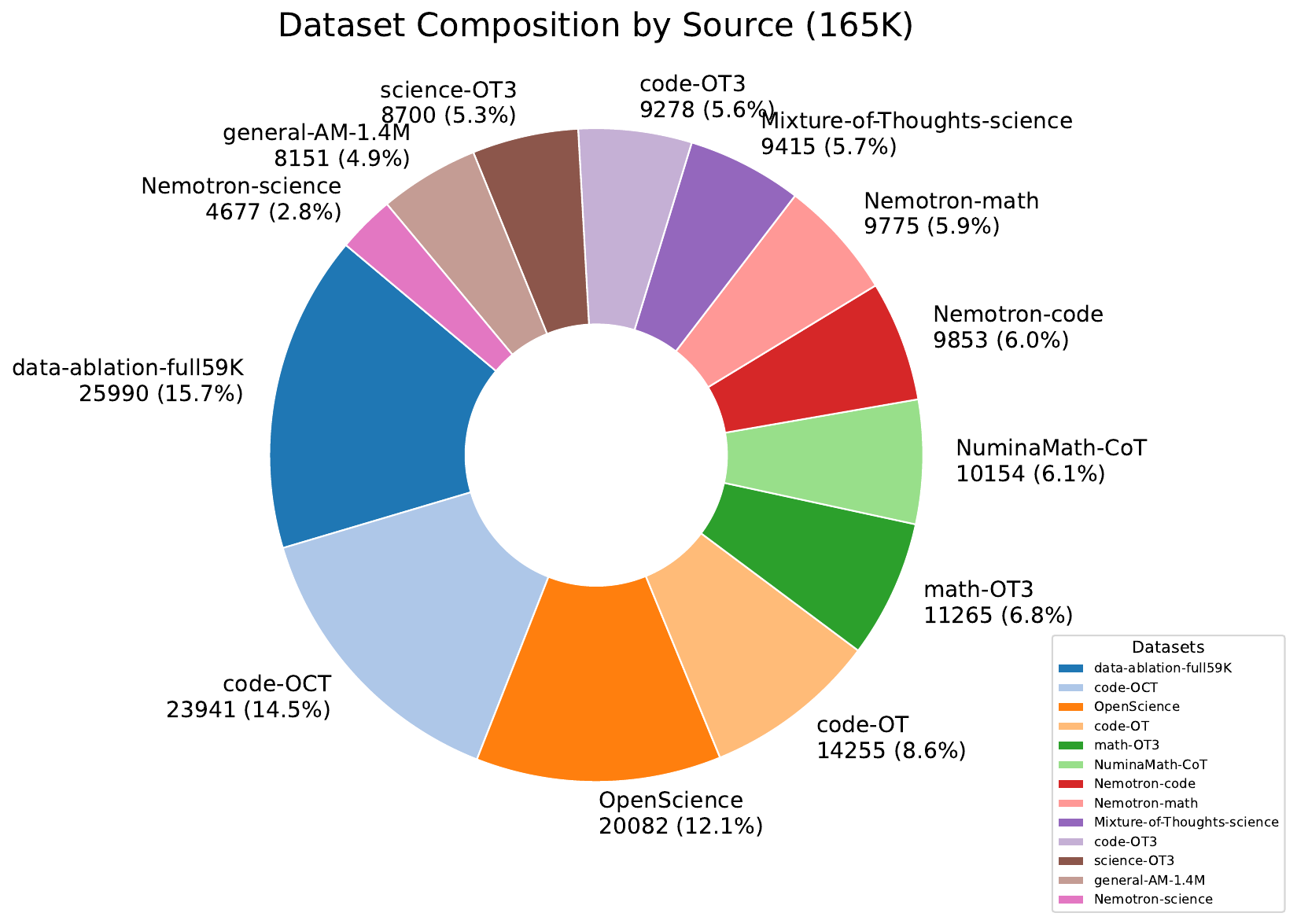}
\caption{Data Composition and Proportion of full 165K SFT data.} 
\label{main}
\end{figure*}
\twocolumn

\end{document}